\documentclass[journal]{IEEEtranTIE}
\usepackage{graphicx}
\usepackage{cite}
\usepackage{picinpar}
\usepackage{amsmath}
\usepackage{url}
\usepackage{flushend}
\usepackage[latin1]{inputenc}
\usepackage{colortbl}
\usepackage{soul}
\usepackage{multirow}
\usepackage{pifont}
\usepackage{color}
\usepackage{alltt}
\usepackage[hidelinks]{hyperref}
\usepackage{enumerate}
\usepackage{siunitx}
\usepackage{breakurl}
\usepackage{epstopdf}
\usepackage{pbox}
\usepackage{pifont}

\usepackage{amssymb}
\usepackage{booktabs}
\usepackage{placeins}
\usepackage{float}
\usepackage{adjustbox}

\def\blfootnote{\xdef\@thefnmark{}\@footnotetext}

\begin{document}
\title{	

Admittance Visuomotor Policy Learning for General-Purpose Contact-Rich Manipulations

}

\author{

Bo Zhou, \textit{Member, IEEE}, Ruixuan Jiao, Yi Li, Xiaogang Yuan, \\ Fang Fang, and Shihua Li,\textit{ Fellow, IEEE}

\thanks{
    This work was supported in part by the National Natural Science 
    Foundation (NNSF) of China under Grant 62073075. \textit{(Corresponding author: Bo Zhou.)}
    
    The authors are with the School of Automation, Southeast University
    and the Key Laboratory of Measurement and Control of CSE, Ministry of Education, Nanjing 210096, China (e-mails: {zhoubo@seu.edu.cn}, {220232086@seu.edu.cn}; {220221974@seu.edu.cn}; {ffang@seu.edu.cn}; \\ {230248976@seu.edu.cn}; {lsh@seu.edu.cn})
    }%
}

\maketitle

\begin{abstract}

Contact force in contact-rich environments is an essential modality for robots to perform general-purpose manipulation tasks, as it provides information to compensate for the deficiencies of visual and proprioceptive data in collision perception, high-precision grasping, and efficient manipulation.
In this paper, we propose an admittance visuomotor policy framework for continuous, general-purpose, contact-rich manipulations.
During demonstrations, we designed a low-cost, user-friendly teleoperation system with contact interaction, aiming to gather compliant robot demonstrations and accelerate the data collection process.
During training and inference, we propose a diffusion-based model to plan action trajectories and desired contact forces from multimodal observation that includes contact force, vision and proprioception. We utilize an admittance controller for compliance action execution. 
A comparative evaluation with two state-of-the-art methods was conducted on five challenging tasks, each focusing on different action primitives, to demonstrate our framework's generalization capabilities. 
Results show our framework achieves the highest success rate and exhibits smoother and more efficient contact compared to other methods, the contact force required to complete each tasks was reduced on average by 48.8\%, and the success rate was increased on average by 15.3\%. 
Videos are available at \textcolor{blue}{\href{https://ryanjiao.github.io/AdmitDiffPolicy/}{https://ryanjiao.github.io/AdmitDiffPolicy/}}.

\end{abstract}


{}

\definecolor{limegreen}{rgb}{0.2, 0.8, 0.2}
\definecolor{forestgreen}{rgb}{0.13, 0.55, 0.13}
\definecolor{greenhtml}{rgb}{0.0, 0.5, 0.0}

\section{Introduction} 
\begin{figure}
    \centering
    \setlength{\abovecaptionskip}{0.cm}
    \includegraphics[width=0.95\linewidth]{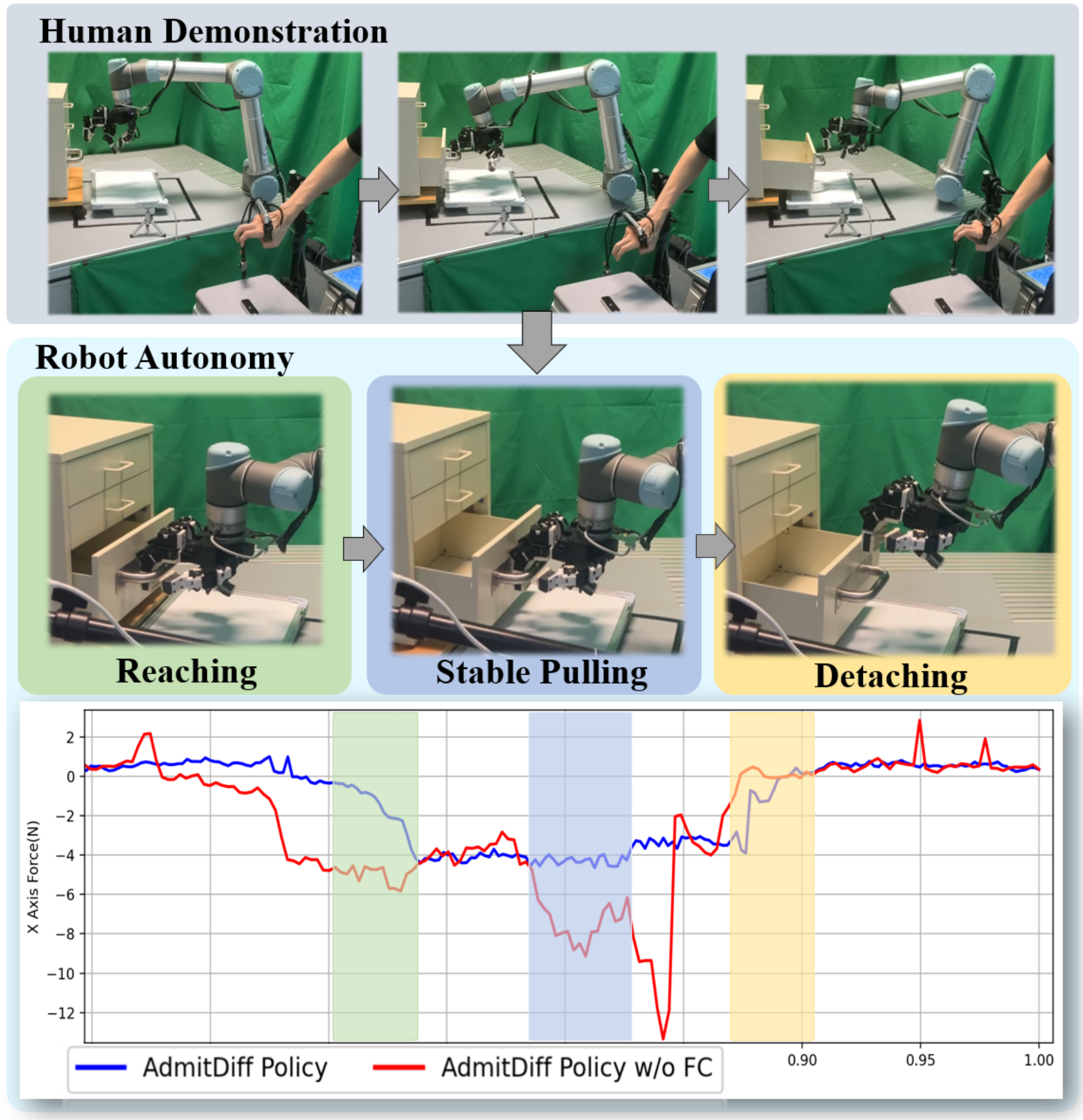}
    \caption{Robot learning multi-stage, contact-rich manipulation skills from human demonstration to open a drawer with AdmitDiff Policy framework. 
    Compared to a policy without the force control module, our completed framework demonstrates a smoother switching between contact stages and significantly reduces the contact force required to complete the task.}
    \vspace{-3mm}
    \label{fig:overview}
\end{figure}

\IEEEPARstart{R}{obots} performing daily manipulation tasks for humans demands the ability to execute a wide variety of actions with precision and adaptability. 
Actions can be classified into types such as grasping, pushing, pulling, rubbing and twisting, most of which involve object contact and require precise force adjustments.
Moreover, manipulations in daily environment are typically divided into multi-tasks and multi-stages, with distinct force control requirements. The stage switching is also particularly challenging to identify. For instance, the action of opening a drawer can be divided into three stages: Reaching, Pulling, and Detaching, handled imprecisely may cause force pikes as illustrated in Fig.~\ref{fig:overview}. 
To address these diverse and intricate challenges, robots require both advanced hardware and sophisticated control policy.
Multi-fingered hand robots have demonstrated effectiveness in performing general-purpose, multi-stage manipulation tasks.
However, deploying multi-fingered hand robots in complex manipulation environments present significant challenges.

The planning and control for multi-fingered hand robots in general-purpose, multi-stage, contact-rich manipulations features high-dimensional observation space and multi-modal action distributions.
Model-based methods, which rely on precise dynamics models and state estimation, are often inadequate in complex manipulations \cite{yang2024imitate}.
Data-driven methods provide the capability to achieve complex manipulation skills. 
These methods typically rely on extensive, high-quality demonstration data, which is crucial for model efficiency and generalization.

Human teleoperation serves as an intuitive and efficient method for collecting demonstrations, considering the structural similarity between these robots and the human hand-arm.
Vision-based robot teleoperation methods, as a mainstream approach for hand-arm teleoperation, offer advantages in lower cost and ease of deployment compared with methods using tracker-based virtual reality devices and wearable gloves \cite{Zhao2020leap, qin2023anyteleop}.
Despite being widely adopted, the quality of demonstrations collected through vision-based teleoperation is difficult to guarantee. 
This method faces challenges in providing interactive feedback to the operator, as well as precise detection of human hand movements and finger articulations. 
Contact force is critical for delivering intuitive feedback on contact states, thereby enhancing telepresence awareness and improving operator responsiveness.
Current approaches that utilize artificial skin and pneumatic gloves for force feedback are expensive and prone to damage.
For contact-rich manipulation data collections that demand high force-position control accuracy, these limitations need to be urgently addressed.

Existing data-driven methods for general-purpose contact-rich manipulations focus on robot trajectory planning, where contact force control causally influences task completion \cite{yue2024visualadmit}.
In multi-stage, contact-rich manipulation tasks, the force-control requirements at different stages must be satisfied, and switching between stages need to be handled smoothly, posing significant challenges for policy design.
To the best of the authors' knowledge, there is limited work addressing continuous force control within general-purpose multi-stages manipulations, particularly regarding the planning and control of multi-fingered hand robots.
Diffusion-based visuomotor policy excels at generating high-dimensional data that aligns with the desired action distributions \cite{chi2023diffusion, prasad2024consistency}, offering significant advantages in planning and control for multi-fingered systems.
However, diffusion-based policy that rely on trajectories and images are generally focused on perception and planning, gives insufficient attention to force control stage transition and control accuracy, greater emphasis should be placed on the integration of vision, trajectory, and contact force.

The main contributions of this paper are as follows:
\begin{enumerate}[1)]

\item  A low-cost, user-friendly teleoperation system is designed for efficient data collection. This system integrates multi-modal perception, in particular utilizing vibration motors to provide contact force feedback, thereby enhancing operator engagement, it also employs dual stereo cameras for precise pose tracking. Notably, high-quality multi-model demonstration data can be obtained, with the total system cost remaining below \$400.

\item A diffusion-based admittance visuomotor policy, termed the AdmitDiff Policy, is proposed for continuous, general-purpose, contact-rich manipulation tasks.
Specifically, our method encodes contact force and employs a multi-head mechanism to predict dynamic contact force and trajectory, enhancing force awareness. An admittance controller was implemented to smoothly execute policy actions.
Fig.~\ref{fig:overview} shows the effectiveness of our policy framework.

\item Five challenging contact-rich multi-stages manipulation tasks are conducted to evaluate the generalization performance of our framework. 
Our framework is compared with other state-of-the-art algorithms and the experimental results indicate that our method outperformed other algorithms, the contact force was reduced on average by 48.8\%, and the success rate was increased on average by 15.3\%. 

\end{enumerate}

Our code and design files will be publicly available \footnote{\href{https://ryanjiao.github.io/AdmitDiffPolicy/}{https://ryanjiao.github.io/AdmitDiffPolicy/}}.

\section{Related Works}
\subsection{Robot Teleoperation}

In general-purpose contact-rich manipulation tasks, industrial robot teach pendants often face difficulties in managing complex task environments and varying task requirements. Recent advancements in sensor technology have led to increased efforts toward enabling dexterous robot teleoperation by capturing human motion through various sensors and corresponding algorithms.

Tracker-based motion capture devices can capture human body movements, and this capability has been applied to robot control. For instance, \cite{zhang2018deep} used virtual reality devices to teleoperate the PR2 robot. The handheld controller captures the hand's position and pose in real-time, while the user's head-mounted display enhances reality, placing the human and robot in the same observational space. These methods with virtual reality devices tend to be costly.\cite{Li2020handarmTele} use Inertial Measurement Unit (IMU) and depth camera to capture human motion for robot control. 

Vision-base methods is widely adopted in robot teleoperation, \cite{qin2023anyteleop, handa2020dexpilot, fu2024humanplus} used tools like MediaPipe to predict hand position, orientation and gesture from image streams, controlling the robot's end-effector and dexterous hand posture. This approach was successfully deployed on various robot platforms, such as UR5, Kuka, and Xarm. These monocular vision-based methods have delays and inaccuracies in estimating wrist poses and hand joint positions. Some researchers have designed specific master-slave control mechanisms for robot arm drag teaching \cite{darvish2023teleoperation, zhao2023learning, wu2023gello}. Other researchers focused on portability \cite{chi2024universal, wang2024dexcap}. They designed handheld grippers similar to robot grippers and used a GoPro camera running Visual-Inertial Odometry (VIO) algorithm to record the operator's hand position and orientation, remapping it to the robot's end-effector orientation. This approach completed tasks like washing dishes and folding clothes. These methods are difficult to adapt to multi-fingered hands. Each of these three approaches has limitations. Our method balances cost and effectiveness, maintaining wrist pose estimation stability at a low cost, enabling hand pose control, and incorporating force feedback functionality.

\subsection{Robot Learning from Demostrations}

Learning from human demonstrations is an effective approach for enabling robots to acquire general manipulation skill. The concept of behavioral cloning has been widely applied in this context \cite{zhao2023learning, wang2023mimicplay, chi2023diffusion, sridhar2024nomad}. 
Researchers have addressed robot action prediction by formulating it as a time series problem, utilizing transformer-based networks to improve prediction accuracy. One study \cite{zhao2023learning} used transformer to predict an entire sequence of future actions, thereby reducing the impact of cumulative errors on robot decision-making and enabling the learning of various manipulation policies in high-dimensional action spaces on a dual-arm platform. 
To address issues such as single-action prediction and the high-dimensionality of action spaces, one study \cite{chi2023diffusion} introduced probabilistic diffusion models into robot action prediction, enhancing the robustness of the algorithm. Additionally, to reduce the computational cost of high-dimensional continuous control, the study predicted the robot's actions for the next 8 steps, thereby lowering computational demands. \cite{sridhar2024nomad} applied diffusion models to goal-oriented navigation problems, predicting the robot's movement direction based on the target scene and current observations. 

Furthermore, \cite{haldar2024baku} have explored encoding human instructions using pre-trained text or visual encoders, embedding the encoded information into visual representations through Feature-wise Linear Modulation (FiLM), thereby enabling robots to perform multiple tasks simultaneously. Another study \cite{yang2023moma} concentrated on the aspect of force interaction within imitation learning. This study sought to enhance the robot's performance in tasks involving intricate force interactions by employing compliant control strategies.

Our approach leverages the advantages of diffusion models in fitting high-dimensional action distributions. We designed a network capable of accurately fitting various distinct actions while simultaneously predicting the robot's end-effector pose, hand joint movements, and contact force. These predicted contact force are then used to correct kinematic errors in real-world tasks, thereby achieving force compliance in complex contact operations. 
To address the control latency associated with the slow inference speed of multi-step denosing diffusion models, we employed a student-teacher distillation \cite{prasad2024consistency}. This approach enables the model to execute single-step denoising for action sequence prediction, thereby establishing improved conditions for responsive admittance control.

\subsection{Force Control in Learning-based Manipulation}

Interactive force control methods are widely employed in manipulation tasks, as discussed in \cite{beltran2020variable, ren2018learning}. In contact-rich environments, the inherent uncertainties often lead to challenges with position-based control strategies, which may not effectively prevent joint and structural damage during tasks requiring significant contact force. To address this, direct and indirect force control strategies, typically incorporating external sensors, offer precise regulation of force through motion control. Among the commonly utilized compliance control approaches are impedance control \cite{hogan1985impedance} and admittance control \cite{siciliano2008springer}, both of which model the robotic arm as an adjustable mass-spring-damper system, thereby enabling dynamic adaptation of contact force.

Compliance control have been integrated into learning-based manipulation tasks. For instance, \cite{beltran2020learning} employed Proportional-Derivative (PD) and admittance control methods to achieve closed-loop control of contact force during both training and inference phases in robotic reinforcement learning, while simultaneously adjusting the end-effector trajectory. In a separate study, \cite{yang2023moma} utilized a prior knowledge base to predict contact force and end-effector poses, with an admittance-based Whole-Body Control (WBC) system being implemented to execute these predictions. Kamijo et al \cite{kamijo2024learning} proposed the Comp-ACT method, which integrates end-effector pose and six-dimensional force/torque sensor data into an imitation learning framework to predict the robot's end-effector pose and the stiffness parameters for the admittance controller. However, this method relies on a manually tuned constant goal wrench during teleoperation and task execution, necessitating stiffness adjustments to accommodate varying desired contact force.

Compliant control with constant force tracking is often inadequate for complex dexterous manipulation tasks due to the numerous contact states involved and the frequent switching between these states. Our approach surpasses the Comp-ACT methodology by estimating dynamic contact force rather than predicting compliant controller parameters. The desired force-position trajectory is interpolated over predicted time steps, while a high-frequency admittance controller dynamically adjusts the motion based on real-time contact force variations, enhancing the robot's ability to handle complex contact-rich tasks.

\section{Methodology}

The AdmitDiff Policy framework predicts the robot movements and contact force by leveraging the robot pose and force feedback. An admittance controller refines robot actions by interpolating the force-position trajectories produced by the policy, ensuring smooth and precise manipulation.

\begin{figure*}
    \centering
    \includegraphics[width=0.95\linewidth]{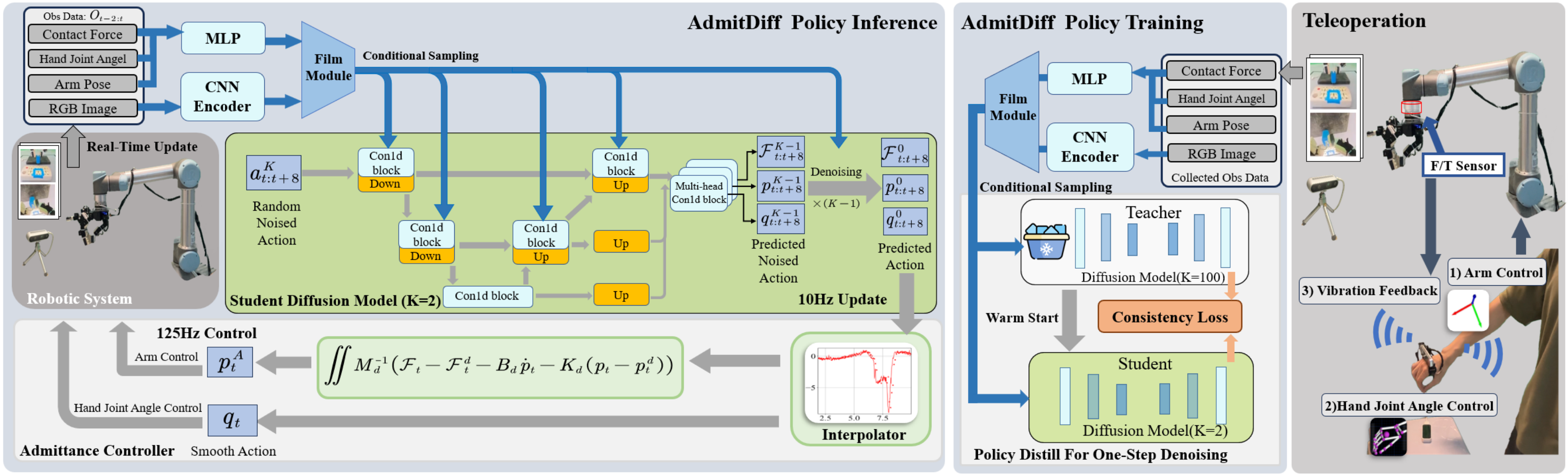}  
    \vspace{-2.5mm}
    \caption{{AdmitDiff Policy Framework.} \textit{Left:} During inference, the previous two steps' observations are encoded as inputs for noise estimation, while the student model outputs actions for the next 8 time steps, the number \(K\) represents the denoising iteration required by the diffuser. The arm's force-position trajectory is used in the admittance controller to compute the desired pose.  \textit{Middle:} The teacher model is trained for 100 denoising steps, then its parameters are frozen to train the student model with a consistency loss for single-step denoising. \textit{Right:}  Data collection, including contact force information, is performed using the teleoperation system designed in this work.}
    \label{fig:policy}
    \vspace{-5mm} 
\end{figure*}

The frameworks for teleoperation, training and inference are illustrated in Fig.~\ref{fig:policy}.

We developed a force-interactive hand-arm robotic system to test our proposed framework and collect human demonstration data, incorporating a cost-effective force feedback teleoperation method that can be adapted to different hand-arm robotic systems. To achieve general manipulation tasks, we chose the LEAP Hand, an open-source hardware by Shaw \cite{shaw2023leap}, as the end-effector. Its simple design, featuring 3D-printed parts and servo assemblies, offers both dexterity and ease of maintenance.

\begin{figure}
    \centering
    \includegraphics[width=1.0\linewidth]{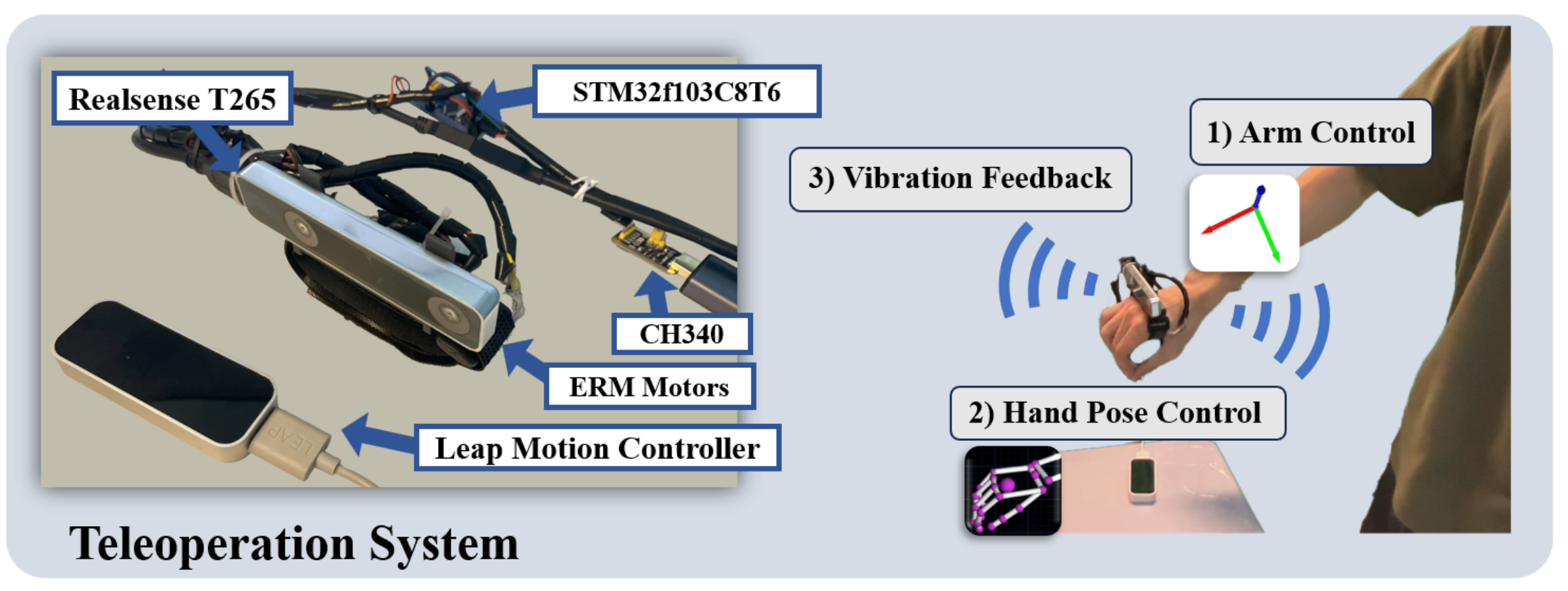}
    \caption{The teleoperation system, worn and used by an operator, is shown in the figure on the right, with the three functional components highlighted. Specific component models are labeled on the left.}
    \vspace{-3mm}
    \label{fig:teleop_system}
\end{figure}

\subsection{Force-Interactive Hand-Arm Teleoperation} 
The teleoperation system is detailed illustrated in Fig.~\ref{fig:teleop_system}. There are three main components provide critical data for the teleoperation task.

\subsubsection{Wrist spatial pose control module} It estimates the 6-DoF position and orientation of the wearer's wrist using the Intel RealSense Tracking Camera T265. The Intel RealSense T265 integrates Stereo cameras and an IMU to run a Visual-Inertial Odometry to provide accurate spatial odometry. Following the approach in \cite{handa2020dexpilot}, we designed a simple glove using Velcro and 3D-printed parts to securely mount the T265 on the operator's palm.

\subsubsection{Gesture detection and remapping module} It uses the Ultraleap Leap Motion Controller to estimate the keypoints coordinate of human hand gestures. 
Then generate the joint angles of multi-fingered hand robot with detected human hand gestures to control the movements of a multi-fingered robotic hand, enabling precise gesture-based interaction with the robot.

\subsubsection{Wrist-mounted force feedback module} It uses Eccentric Rotating Mass (ERM) vibration motors attached to the operator's palm. Our system uses an stm32f103c8t6 microcontroller to generate Pulse Width Modulation (PWM) signals for a NPN Bipolar Junction Transistor, based on force data from the wrist Force/Torque sensor. Contact force between 0-20N are mapped linearly to vibration intensity, providing real-time feedback to the operator, which is similar to \cite{ding2024bunny}.

During the teleoperation process, operator controls the robotic arm with T265 tracking camera, and can adjusts their movements based on vibration feedback from the ERM vibration motors. This could control the contact force within an acceptable range to achieve force compliance. 
Meanwhile, operator controls the robotic hand with Leap Motion Controller. It could capture 21 spatial coordinate keypoints from human hand images to detect hand gestures. A kinematic retargeting process is used to map the human hand gestures onto the robotic hand, as outlined in \cite{qin2023anyteleop, handa2020dexpilot}. Task-space keypoints vectors are formulated from hand spatial coordinate points to evaluate the loss of mapping. An optimization problem is introduced to minimize the vector error within the task space, with the cost function defined as follows:

\vspace{-7pt}
\begin{equation}
\begin{aligned}
\mathit{C}(v_t^i, q_t) &= \sum_{i=0}^{N} \left( \left| \alpha v_t^i - f_i(q_t) \right|^2 \right) + \beta \left| q_t - q_{t-1} \right|^2    \\
\text {s.t. } & q_l \leq q_t \leq q_u 
\end{aligned}
\end{equation}

where \(v_t^i\) is the keypoint vector of the human hand, \(f_i(q_t)\) is the keypoint vector calculated based on the joint angles \(q_t\) of the robotic hand, and \(\alpha\) and \(\beta\) are weight coefficients. 

\subsection{AdmitDiff Policy Design and Training} 

Diffusion models have already been successfully applied in robotic action generation and have demonstrated advanced performance in the field of imitation learning.

\subsubsection{Policy Design} Our method incorporates proprioceptive, vision and force modalities via a diffusion-based imitation learning policy.
The policy predicts future desired end-effector positions, orientations, and contact force. This mapping relationship can be expressed as:
\vspace{-3mm}

\begin{equation}
\begin{aligned}
\vspace{3mm}
f(I_{\text{global}}, I_{\text{local}}, \mathit{s}_{\text{proprio}}, \mathcal{F}_{\text{current}})  
\rightarrow (\mathit{p}_{\text{ee}}, \mathit{q}_{\text{\ finger}}, \mathcal{F}_{\text{desired}})  
\end{aligned}
\end{equation}

In this model, global and local camera images \(I_{\text{global}}, I_{\text{local}}\), proprioceptive data \(\mathbf{s}_{\text{proprio}}\), and contact force measurements from a wrist Force/Torque sensor \(\mathcal{F}_{\text{current}}\) serve as input features. These inputs 
are linearly projected into a unified feature space via a FiLM module, which conditions the diffusion model to facilitate accurate noise prediction. Our noise prediction network is structured similarly to the diffusion policy.
We employ ResNet-18 as the CNN encoder and utilize a 3-layer MLP as the low-dimensional data encoder. 
However, to mitigate information loss during the decoding phase and enhance the precision of action predictions, we extract features from multiple levels during the upsampling process, normalizing them to the same dimension and subsequently fusing them for final action prediction. 
To ensure that different types of action-related information do not interfere with each other, we employ multiple prediction heads, each Multi-Head Conv-Block consists of four convolutional layers designed to gradually reduce the number of input channels and extract distinct features for conditional denoising, allowing for a more precise formulation of the loss function.

\subsubsection{Policy Training} While diffusion models are highly effective for generating multimodal robotic actions, the iterative multi-step denoising process can be computationally intensive, potentially compromising the compliance and dynamic performance of subsequent admittance control. Drawing inspiration from the Consistency Model (CM)\cite{prasad2024consistency}, we introduce a consistency distillation approach into our AdmitDiff Policy framework. By applying this method within the Probability Flow Ordinary Differential Equation (PFODE) framework, we train the model to produce consistent predictions across various points along the trajectory, ultimately enabling single-step denoising. The improvement in inference time from single-step denoising is mentioned in Section~\ref{subsec:Setups}.

The teacher model \({\pi}_{\theta}^{T}\) is trained within the EDM \cite{karras2022elucidating} framework:

\begin{equation}
\begin{aligned}
\frac{dA_t}{dt} = \frac{\left[ A_t - {\pi}_{\theta}^{T}(A_t, t; \text{obs}) \right]}{t} 
\end{aligned}
\end{equation}

Its loss function is:

\begin{equation}
\begin{aligned} 
\mathcal{L}_{DSM} = \mathbb{E}&_{\text{batch}} \left[\rho\left( A_0, {\pi}_{\theta}^{T} (A_t, t; \text{obs})\right)\right] \\[4pt]
\rho(x, y) &= \sqrt{||x - y||^2 + c^2} - c 
\end{aligned} \label{eq
} \end{equation}

where \(c = 0.005 \sqrt{D}\), D is dimension of action. Given the PFODE trajectory of \(A_t\), the initial \(A_0\) can be estimated, but at this time, the teacher model does not yet have the ability to predict \(A_0\) accurately in a single step. To achieve single-step denoising, the teacher model \({\pi}_{\theta}^{T}(A_t, t; \text{obs})\) needs to be distilled into the student model \({\pi}_{\phi}^{S}(A_t, t, t'; \text{obs})\). The student can obtain \(A_{t'}\) at any earlier time from the action \(A_t\) at time \(t\). 

The student model, initialized with the teacher model, is optimized with the goal of making predictions \(s\) at any two points \(u, v\) on the PFODE trajectory as close as possible:

\begin{equation}
\begin{aligned}
\mathcal{L}_{CTM} = \mathbb{E}_{\text{batch}} [\rho(& {{\pi}_{\phi}^{S}} (A_v, v, s; \text{obs}), \\
& {{\pi}_{\phi}^{S}} (A_u, u, s; \text{obs}))]
\end{aligned}
\end{equation}

where \(s < u < v\), \(A_u = {{\pi}_{\theta}^{T}} (A_v, v, u; \text{obs})\). The consistency loss \( \mathcal{L}_{\textit{cons}} \) for distilling the student model is:

\begin{equation}
\begin{aligned}
 \mathcal{L}_{\textit{cons}} =  k_0 \mathcal{L}_{CTM} + k_1 \mathcal{L}_{DSM}
\end{aligned}
\end{equation}

The trajectory points produced by the Diffusion Policy and Consistency Policy are updated at a frequency of 10Hz. In order to achieve a smooth and continuous of force control, it is essential to interpolate the force-position trajectory points to achieve finer temporal resolution. 

\subsection{Admittance Force Control}
Firstly, in robotic manipulation tasks, we define some common physical quantities. The pose of the robot's end-effector is represented by {\small \( \mathcal{P}(t) = [R(t), p(t)] \in SE(3) \)}, 
where {\small \( R(t) \in SO(3) \)} and {\small $p(t) = [x(t), y(t), z(t)]^T  \in \mathbb{R}^3$} represent the rotation and position of the end-effector, respectively.
The contact force vector in cartesian space between the robot's end-effector and the environment is described by {\small $\mathcal{F}(t) = [f_x(t), f_y(t), f_z(t)]^T \in \mathbb{R}^3$}.

Secondly, we utilize a admittance model presented in \cite{liu2007fuzzy} to control the contact force between the robot end-effector and an unknown external environment when contact is necessary to complete the task:

\begin{equation}
\begin{aligned}
M_d (\ddot{p}(t) &- \ddot{p}_d(t)) + B_d (\dot{p}(t) - \dot{p}_d(t))\\ 
& + K_d \left( p(t) - p_d(t) \right)  
= \mathcal{F}(t) - \mathcal{F}_d(t)  \label{equ:7}
\end{aligned}
\end{equation}

where \( M_d, B_d, K_d \in \mathbb{R}^{3 \times 3} \) denote the inertia, damping, and stiffness matrices during the contact between robot end-effector and environment, which regulate the contact behavior. 

\( \mathcal{F} \) and \( \mathcal{F}_d \) represent the feedback and desired force vectors of the robot end-effector. \( p \) and \( p_d \) represent the feedback and desired positions of the robot end-effector. According to equation.~\ref{equ:7}, the acceleration can be derived as:

\begin{equation}
\begin{aligned}
\ddot{p}(t) = M_d^{-1} ( &\mathcal{F}(t) - \mathcal{F}_d(t) - B_d \dot{p}(t) \\
                              & - K_d ( p(t) - p_d(t) ) )
\end{aligned}
\end{equation}

Then we can update the acceleration and position according to the dynamics. The position will be used to control the robotic arm during inference:

\begin{equation}
\begin{aligned}
\dot{p}(t + \Delta t) = \dot{p}(t) + \ddot{p}(t) \cdot \Delta t \\
p(t + \Delta t) = p(t) + \dot{p}(t) \cdot \Delta t
\end{aligned}
\end{equation}

AdmitDiff Policy estimates a consistent future force-position trajectory points, with each point spaced 0.1 seconds apart. Before the admittance controller takes these trajectory points as desired inputs, we perform interpolation to generate a dense trajectory curve. Specifically, we use one-dimensional linear interpolation (Interp1d) for the target positions and contact force, and spherical linear interpolation (Slerp) for the target rotations. This ensures smooth and continuous trajectories for the controller to follow. We manually design the corresponding \( M_d \), \( B_d \), and \( K_d \) parameters for different tasks to achieve better force tracking performance.

\vspace{5pt}

\section{Evaluation on Robot Teleoperation}
\subsection{System Setup}
We have developed an integrated system for data collection and algorithm validation, which can be separated into the slave-side and master-side.

\subsubsection{Slave-Side System}
The slave-side system includes a 6-DoF UR5-CB3 robotic arm with a LEAP Hand as the end-effector. An OptoForce 6-Axis Force/Torque Sensor is mounted on the wrist for force feedback control, and visual feedback is provided by a RealSense D435i camera and a UVC camera on the LEAP Hand. 

\subsubsection{Master-Side System}
The master-side system consists of an Intel RealSense T265 tracking camera, a Leap Motion Controller, an STM32 microcontroller, and ERM vibration motors.

The entire system is controlled by a computer platform with Intel i9-12900K processor and NVIDIA RTX 3090 GPU. The policy training and inference is also deployed on this platform.

\subsection{Performance of Robot Teleoperation}

\begin{table}[h]
    \centering
    \vspace{-2mm}
    \scriptsize
    \setlength{\tabcolsep}{11pt} 
    \caption{\raggedright Teleoperation Systems Evaluation Results in Telekinesis Benchmark.}
    \label{tab:teleop_success_rate}
    \begin{adjustbox}{max width=0.5\textwidth}
    \begin{tabular}{l l c c}
    \toprule
    \textbf{Task} & \textbf{System} & \textbf{Success} & \textbf{Time (s)} \\
    \midrule
    \multirow{4}{*}{Pickup Box Object} 
    & SpaceMouse+Leap  & 17/20 & 20.7  \\
    & AnyTeleop  & 20/20 & 16.2 \\
    & Ours w/o FF & 19/20 & 14.5 \\
    & \textbf{Ours} & \textbf{20/20} & \textbf{11.1}  \\
    \midrule
    \multirow{4}{*}{Pickup Fabric Toy} 
    & SpaceMouse+Leap  & 12/20 & 21.3  \\
    & AnyTeleop  & 17/20 & 17.0 \\
    & Ours w/o FF & 18/20 & 15.8  \\
    & \textbf{Ours} & \textbf{18/20} & \textbf{13.7}  \\
    \midrule
    \multirow{4}{*}{Box Rotation} 
    & SpaceMouse+Leap  & 15/20 & 24.5  \\
    & AnyTeleop  & 18/20 & 18.5  \\
    & Ours w/o FF & 19/20 & 17.0 \\
    & \textbf{Ours} & \textbf{20/20} & \textbf{13.9} \\
    \midrule
    \multirow{4}{*}{Open Drawer} 
    & SpaceMouse+Leap  & 12/20 & 21.5  \\
    & AnyTeleop  & 16/20 & 20.2  \\
    & Ours w/o FF & 18/20 & 16.0  \\
    & \textbf{Ours} & \textbf{20/20} & \textbf{13.3} \\
    \midrule
    \multirow{4}{*}{Two Cup Stacking} 
    & SpaceMouse+Leap  & 14/20 & 24.5  \\
    & AnyTeleop  & 14/20 & 20.0 \\
    & Ours w/o FF & 16/20 & 18.5  \\
    & \textbf{Ours} & \textbf{18/20} & \textbf{15.8}  \\
    \bottomrule
    \vspace{-8mm} 
    \end{tabular}
    \end{adjustbox}
\end{table}

\begin{figure*}[t]
    \centering
    \includegraphics[width=0.9\linewidth]{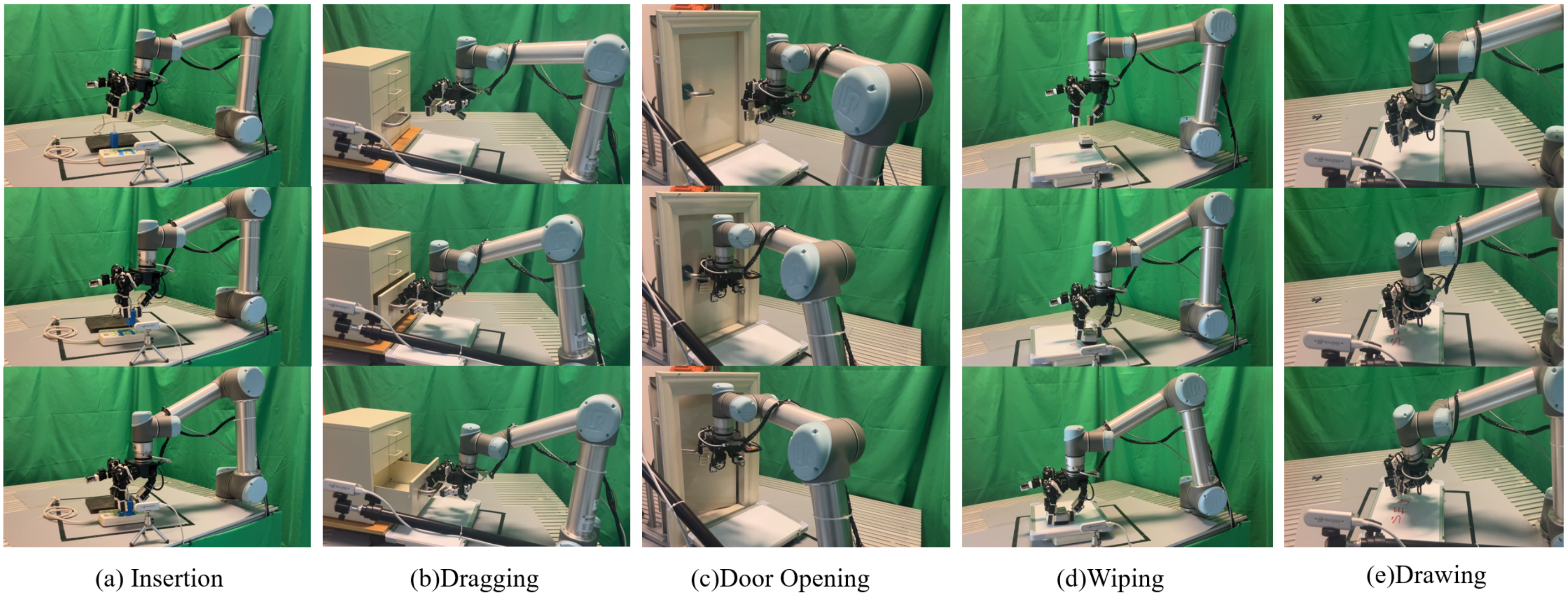}
    \caption{{Task Process Summary Diagram.} We designed five contact-rich manipulation tasks to evaluate the effectiveness of AdmitDiff Policy, and compared them with other imitation learning algorithms. The success rate and contact force optimization results for different tasks are discussed later in this paper.}
    \vspace{-3mm} 
    \label{fig:all_the_task}
\end{figure*}

In order to ensure efficient deployment in contact-rich manipulation tasks, the effectiveness of the entire system in performing complex operations needs to be validated.
We tested our robot teleoperation system on the Telekinesis Benchmark \cite{qin2023anyteleop} to evaluates the performance of multi-finger robotic teleoperation in various tasks. Methods without force feedback (w/o FF) was also evaluated for better comparison. 

Each method was tested 20 times per task. The evaluation results can be found in Table.~\ref{tab:teleop_success_rate}.  Our approach exhibited superior precision and speed over monocular vision-based systems (e.g. Telekinesis, AnyTeleop), while also offering a better user experience than non-wearable devices (e.g. SpaceMouse). 
On average, our method reduced task time by 8.94 seconds compared to the SpaceMouse+Leap system, and by 4.82 seconds compared to AnyTeleop. Additionally, it increased the success rate by 26\% over SpaceMouse+Leap, and by 11\% over AnyTeleop.

\section{Experiments}

\subsection{Experiment Setups} \label{subsec:Setups}

We designed five contact-rich manipulation tasks as the experiment. Each task represents a specific action primitive. The schematic of the experiment is shown in Fig.~\ref{fig:all_the_task}. The detailed task setups are as follows:

\subsubsection{Insertion(grasping and twisting)} The robot grasps a randomly positioned charger plug, stabilizes it with three fingers, and aligns it with a 2 cm hole marked with blue tape. Misalignment is corrected by rotating the plug and adjusting force along the X and Y axes. Visual feedback is from a global and palm-mounted camera. Admittance controller: \(M_d = 3\), \(B_d = 270\), \(K_d = 1540\).

\subsubsection{Door Opening(pushing)} The robot applies 20-30 N of force to open a door, using a downward palm force to rotate the handle, then pushing along the X-axis. Visual feedback is from a global camera. Admittance controller: \(M_d = 10\), \(B_d = 900\), \(K_d = 5000\).

\subsubsection{Dragging(pulling)} The robot pulls a drawer with a randomized position. It inserts two fingers into the drawer handle and applies a consistent force along the X-axis. Visual feedback is provided by a global camera. Admittance controller: \(M_d = 6\), \(B_d = 550\), \(K_d = 3000\).

\subsubsection{Wiping(rubbing)}  This task requires the robot to wipe a whiteboard. The robot grasps a randomly placed eraser with three fingers, moves to the marked area, and applies force along the Z and X axes to clean.  Visual feedback is from both a global and palm-mounted camera. Admittance controller: \(M_d = 5\), \(B_d = 300\), \(K_d = 2000\).

\subsubsection{Drawing(scratching)} The robot grasps a marker and maintains consistent force along the Z-axis while following a trajectory. Visual feedback comes from a global and palm-mounted camera. Admittance controller: \(M_d = 3\), \(B_d = 270\), \(K_d = 1540\).

We collected 50 demonstration trajectories per task using our teleoperation system, followed by pre-processing. The robot's end-effector pose, initially represented by a rotation vector, was converted to the Rotation6D format \cite{zhou2019continuity}, and a first-order low-pass filter was applied to the contact force data. Each task was trained over 400 epochs using the 50 demonstrations, taking 8 hours. In the physical system, the inference time for the Diffusion Policy (DP) is around 100 ms, while the Consistency Model (CM) achieves an inference time of approximately 7 ms.

\subsection{Experiment Results} \label{subsec:ExpRes}
\vspace{-2mm}
\begin{table}[H] 
    \caption[Success Rates]{\raggedright Success rates of real experiments on five different tasks. }
    \label{tab:success_rate}
    \begin{adjustbox}{max width=\linewidth} 
    \begin{tabular}{lccccc|c}
    \toprule
    \textbf{Method} & \textbf{Insertion} & \textbf{DoorOpening} & \textbf{Dragging} & \textbf{Wiping} & \textbf{Drawing} & \textbf{Average} \\ 
    \midrule
    Diffusion policy       & 60\%  & 70\% & 80\% & 70\%  & 20\% & 60\% \\ 
    Consistency policy     & 70\% & 75\% & 80\% & 75\%  & 25\% & 70\% \\ 
    Admit policy w/o FC     & 75\% & 80\% & 85\% & 85\% & 40\% & 73\% \\ 
    \midrule
    Admit policy (Ours)     & \textbf{90\%} & \textbf{95\%} & \textbf{95\%} & \textbf{90\%} & \textbf{45\%} & \textbf{83\%} \\ 
    \bottomrule
    \end{tabular}
    \end{adjustbox}
    \vspace{-3mm} 
\end{table}

\subsubsection{Success Rate on Manipulation Tasks} In this section, we gathered result from the experiments on five challenging tasks. We compared our approach with state-of-the-art visuomotor policies, Diffusion Policy and Consistency Policy. Additionally, we removed the force control (w/o FC) module from the Admit Policy framework while keeping the contact force observation encoding to conduct ablation experiments to assess the contribution of our force control features and network components

The success rates of these methods, with each method undergoing twenty trials, are detailed in Table~\ref{tab:success_rate}. 
Compared to the other three methods, our approach achieved an average success rate increase of 15.3\%. In tasks requiring high precision in force control, such as insertion, our method improved the success rate by up to 30\%. In the Drawing task, the high demand for spatial understanding led to low success rates across all methods. However, our approach maintained a competitive advantage due to its enhanced precision in force control.
\begin{figure}[H]
    \centering
    \includegraphics[width=0.9\linewidth]{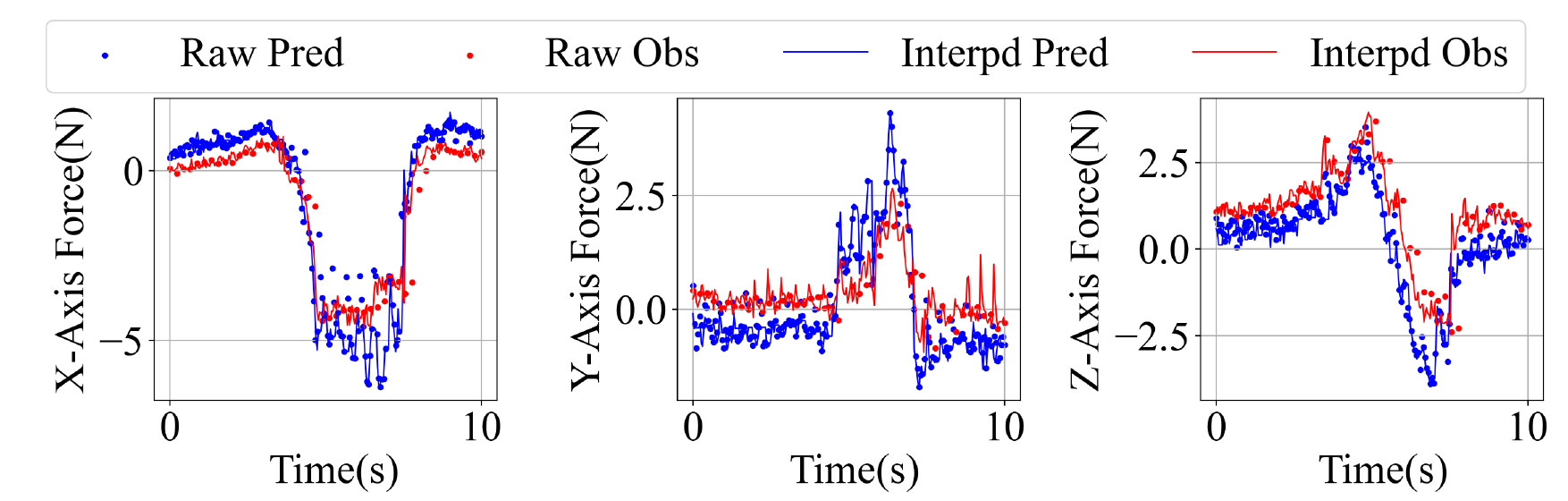}
    \caption{Force control interpolation and tracking performance on different axis in Dragging task.}
    \label{fig:Tracking}
    \vspace{-3mm} 
\end{figure}

 \subsubsection{Performance of Contact Force Tracking} \label{subsubsec:verification} The accuracy of force tracking is crucial for the reliability of policy inference. To demonstrate the interpolation process and force trajectory tracking metrics, we collected data from the Dragging task, as shown in Fig.~\ref{fig:Tracking}.

 The contact force remained stable throughout, maintaining a consistent value of approximately -4.5 N along the negative X-axis. The tracking RMSE was around 0.8 N, allowing for stable and compliant task execution, despite minor fluctuations in the force estimation.
\vspace{-2mm}
 \begin{figure}[h]
    \centering
    \includegraphics[width=0.9\linewidth]{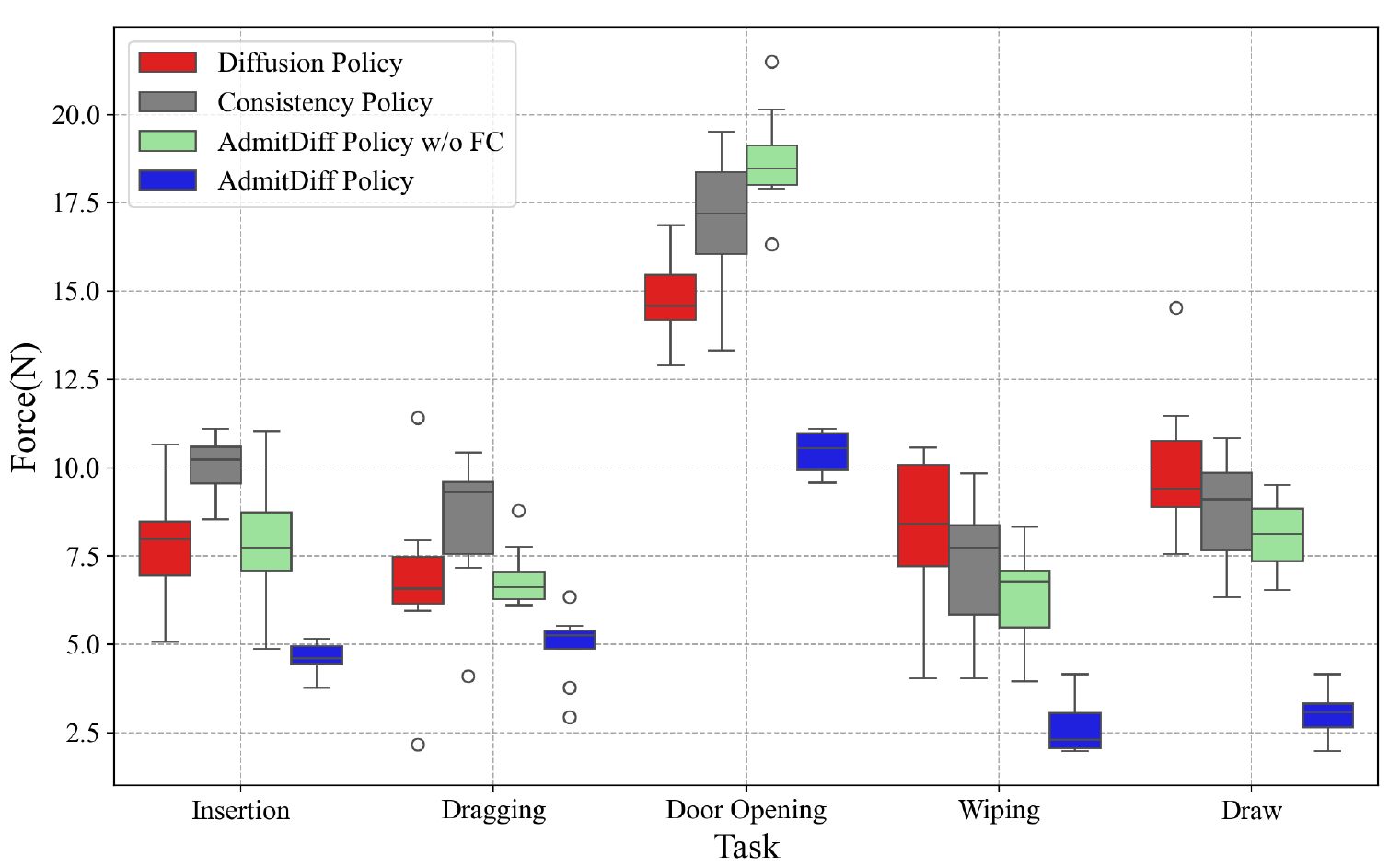}
    \vspace{-2mm} 
    \caption{Comparison of the Mean and Standard Deviation of Contact Force on Five Tasks Using Four Different Methods}
    \label{fig:boxplot}
    \vspace{-3.5mm} 
\end{figure}

\subsubsection{Performance Comparison of Contact Force Control} Fig.~\ref{fig:boxplot} show that AdmitDiff Policy consistently demonstrates the most stable contact force control, with lower variance and more controlled median force, especially in tasks like Insertion and Wiping that require precision. In contrast, Diffusion Policy and Consistency Policy exhibit higher variance, particularly in force-intensive tasks like DoorOpening, indicating less stable control. 

The AdmitDiff Policy shows an average reduction of approximately 48.8\% in mean contact force and 52.0\% in standard deviation compared to the other methods in the five tasks.
\vspace{-4mm}

\begin{figure}[h]
    \centering
    \includegraphics[width=0.85\linewidth]{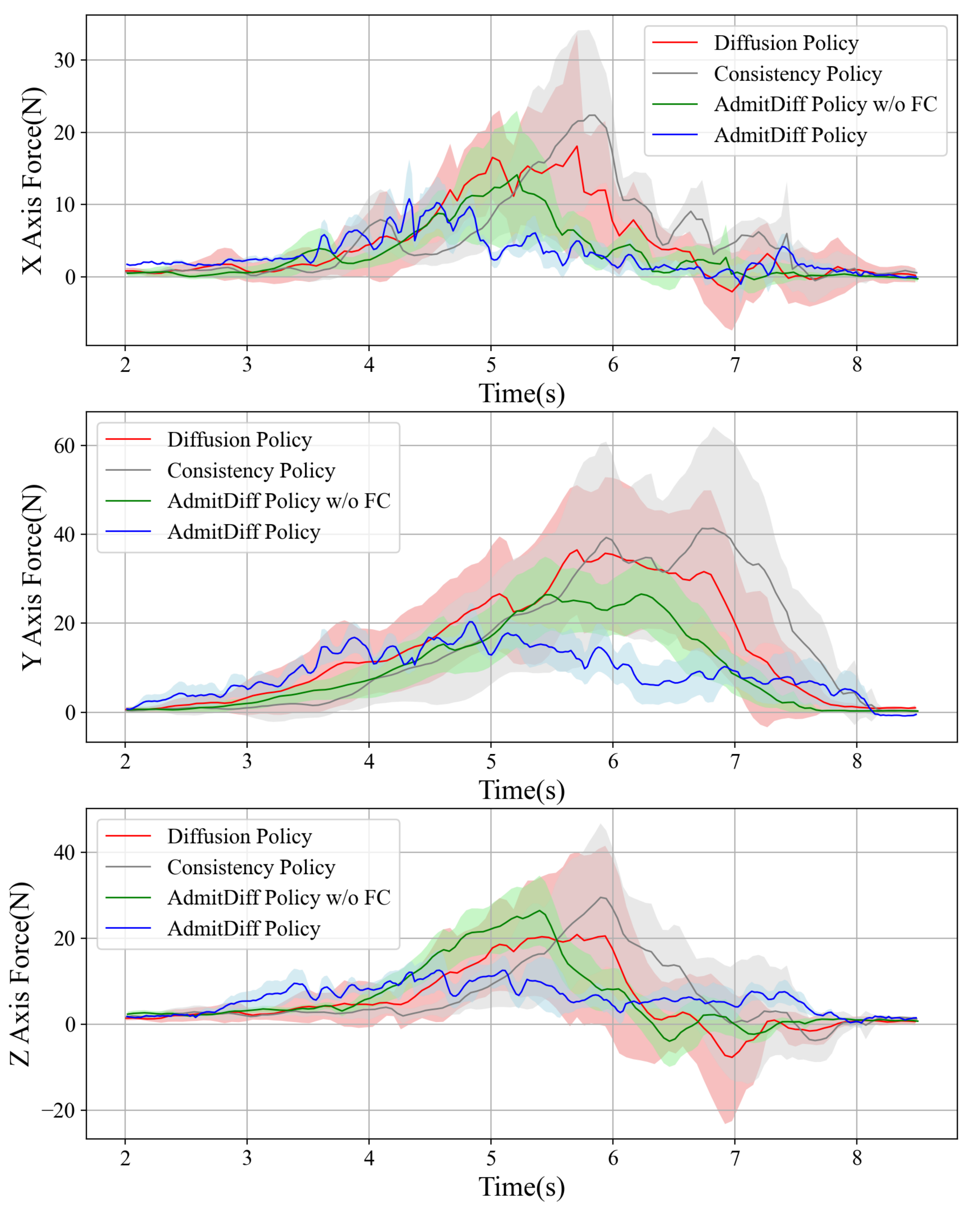}
    \vspace{-3mm} 
    \caption{The Variation in the Mean and Variance of Contact Force During the Entire DoorOpening Task.}
    \label{fig:DoorOpenForceAnalysis}
    \vspace{-3mm} 
\end{figure}

During the DoorOpening task. The result can be seen in Fig.~\ref{fig:DoorOpenForceAnalysis}.
The Diffusion Policy and Consistency Policy show similar performance in mean contact force and contact force variation metrics throughout the task. Compared to the Diffusion Policy, the Admit Policy with the contact force control module reduced mean contact force by up to 53.92\% and force variation by 76.51\%. Even without the control module, adding force feedback reduced force variation by up to 49.21\%. The Consistency Policy, with fewer parameters and faster inference time, showed slight action jitters, leading to larger contact force fluctuations than the Diffusion Policy. The AdmitDiff Policy, using force feedback and an admittance controller, minimized action jitters and significantly reduced the mean contact force and force fluctuations, achieving efficient force control.

\section{Conclusions}
In this paper, we propose a general-purpose admittance visuomotor policy framework for contact-rich manipulation tasks, aimed at improving both success rates and contact force control performance. Additionally, we design a low-cost force-interactive hand-arm teleoperation system to accelerate data collection while ensuring data quality.

This framework significantly enhances both the precision of contact force control and the success rates of tasks, demonstrating its general effectiveness across six distinct actions. Moreover, the evaluation of the teleoperation system showcases highly efficient, high-quality data collection and robust control for contact-rich manipulation tasks.

Future work will focus on extending the application of this method to more complex tasks to further validate its generalization capability and exploring finger-level admittance control to achieve more precise force control operations.

\bibliographystyle{IEEEtranTIE}
\bibliography{ref}

\end{document}